\pdfoutput=1
\documentclass[11pt]{article}
\usepackage{graphicx}
\usepackage[]{acl}
\usepackage{xcolor}
\usepackage{tabularx}
\usepackage{listings}
\usepackage{amssymb}

\definecolor{light-gray}{gray}{0.95} 
\usepackage{tcolorbox}
\usepackage{stfloats}
\usepackage{float}
\usepackage{amsmath}
\newtcolorbox[list inside=prompt,auto counter,number within=section]{prompt}[1][]{
    colbacktitle=yellow!30,
    coltitle=black,
    fontupper=\footnotesize,
    boxsep=5pt,
    left=0pt,
    right=0pt,
    top=0pt,
    bottom=0pt,
    boxrule=1pt,
    title={#1},
    #1, 
}

\lstdefinestyle{mystyle}{
    basicstyle=\ttfamily,
    breaklines=true,
    showstringspaces=false,
    columns=fullflexible,
    commentstyle=\color{olive}, 
    resetmargins=true,
    keywordstyle=\color{blue},
    numberstyle=\tiny\color{gray},
    numbers=none,
    frame=single,
    rulecolor=\color{light-gray}, 
    backgroundcolor=\color{yellow!50}, 
    xleftmargin=5pt, 
    xrightmargin=5pt,
    morestring=[b]",
    morecomment=[l]{//},
    aboveskip=2pt, 
    belowskip=2pt, 
    lineskip=-0pt,
    breakatwhitespace=true,
    breakindent=0em,
    tabsize=4, 
    linewidth=\linewidth, 
    escapeinside={<@}{@>}
}

\usepackage{times}
\usepackage{latexsym}
\usepackage{listings}

\usepackage[T1]{fontenc}

\usepackage[utf8]{inputenc}

\usepackage{microtype}

\title{How Do Humans Write Code? Large Models Do It the Same Way Too}

\author{Long Li$^{1}$\quad Xuzheng He$^{2}$\quad Haozhe Wang$^{3}$\quad Linlin Wang$^{1}$\thanks{\; Corresponding Author}\quad Liang He$^{1}$\\
  $^{1}$East China Normal University, China \\
  $^{2}$Central Conservatory of Music, China\\
  $^{3}$INF Technology, Shanghai, China\\
  $^{1}$\texttt{longli@stu.ecnu.edu.cn} \\
  $^{2}$\texttt{21sa026@mail.ccom.edu.cn} \\}
  
\begin{document}
\maketitle
\begin{abstract}
Program-of-Thought (PoT) replaces natural language-based Chain-of-Thought (CoT) as the most popular method in Large Language Models (LLMs) mathematical reasoning tasks by utilizing external tool calls to circumvent computational errors. However, our evaluation of the GPT-4 and Llama series reveals that using PoT introduces more reasoning errors, such as incorrect formulas or flawed logic, compared to CoT. To address this issue, we propose Human-Think Language (HTL), which leverages a suite of strategies that help integrate PoT and CoT, encompassing: (1) a new generation paradigm that uses full CoT reasoning to control code generation. (2) Focus Attention, that directs model attention to the CoT reasoning during PoT to generate more logical code. (3) reinforcement learning that utilizes the accuracy of both CoT and PoT responses as rewards to prevent repetitive reasoning steps in LLMs when solving difficult math problems. Our method achieves an average improvement of 6.5\% on the Llama-Base model and 4.3\% on the Mistral-Base model across 8 mathematical calculation datasets. It also shows significant effectiveness on five out-of-domain datasets by controlling the model's information flow, exhibiting strong transferability. Additionally, HTL shows the most significant improvement in non-mathematical natural language inference task, contributing to a unified reasoning task framework\footnote{Code is available at: \href{https://github.com/seamoke/Human-Think-Language}{https://github.com/seamoke/Human-Think-Language}}.

\end{abstract}

\section{Introduction}
Solving Mathematical reasoning problems is a significant challenge for current LLMs~\cite{madaan-etal-2022-language,openai2023gpt4}. This task requires interpreting information, identifying relevant mathematical concepts, and formulating equations to solve the problems~\cite{ahn2024large}. Due to computational errors in LLMs~\cite{wei2023chainofthought,gao2023pal}, using CoT~\cite{wang-etal-2023-plan,wei2022chain,chen2024diahaludialoguelevelhallucinationevaluation} solely implemented in natural language can lead to calculation mistakes~\cite{lewkowycz2022solving,wei2023chainofthought,gao2023pal}. The most common practice currently is to use PoT~\cite{chen2023program} for handling mathematical reasoning tasks, by guiding the large model to write the code that is then computed using tool calls. 
\begin{figure}[t]
\centering
\includegraphics[width=1.0\linewidth]{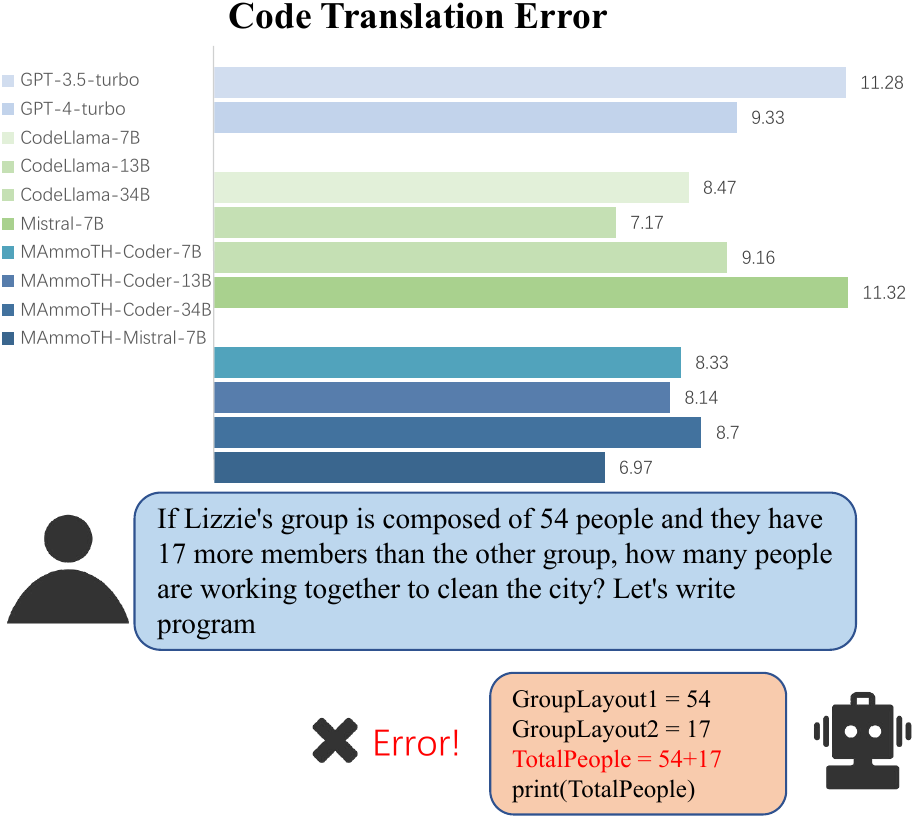}
\caption{The top section of the chart represents the average CTE for each model across 5 datasets. Below is a real example from the Asdiv dataset using the MAmmoTH-Mistral-7B model, which achieved an accuracy of 93.9\% on this dataset. The proportion of CTE remains high across various models, and these errors do not diminish with an increase in model parameters.}\label{fig1}
\end{figure}

However, we made a surprising discovery recently: when a problem is phrased in a manner closer to verbal scenarios (for example, the question is ``One apple costs three dollars, how much for three apples?'' instead of ``3$\times$3=?''), PoT tends to make more reasoning errors or text comprehension mistakes, but this phenomenon is almost non-existent in CoT. For such problems, CoT can correctly reason out the answer, whereas PoT makes mistakes. We refer to this type of error as~\textbf{Code Translation Error (CTE)}. We report the percentage of CTE on five datasets with multiple types of models, the results illustrated in Figure~\ref{fig1}. This error is due to the amount of training data for natural language far exceeding that for code. In the scope of CodeLlama’s pretraining data, which includes 500 billion code tokens, this represents a small fraction compared to the 2 trillion natural language tokens used in the Llama-2 model~\cite{rozière2023code,touvron2023Llama}. Natural language is more suitable for semantic analysis, planning, and abstract reasoning than code~\cite{gou2023tora}.

Existing work also finds the advantage of Natural language, but they have not effectively utilized the reasoning capabilities of natural language. Current research focuses on the following approaches to integrate natural language to enhance the precision of code: (1) Using natural language prompts to guide the model in writing code~\cite{gao2023pal,toshniwal2024openmathinstruct,wang2023mathcoder}: write a brief step in natural language before generating code. (2) Employing methods like self-correction and hybrid approaches to generate answers in multiple stages~\cite{gou2023tora,yue2023MAmmoTH,gou2023critic}. (3) Utilizing prompts like ``rethink question''~\cite{deng2023rephrase} to have the model first paraphrase the question, thereby avoiding comprehension errors.
However, existing methods fall short in two main aspects: First, using few natural language steps or simple paraphrasing methods alone is insufficient for effectively controlling code generation; a more comprehensive natural language reasoning process is necessary to generate more reliable code. Secondly, reasoning within LLMs is not always faithful~\cite{lanham2023measuring,bao2024llms,turpin2023language}. Frequently, the final answers seem to be derived directly from the questions themselves rather than aligning with the reasoning process. Consequently, even correct reasoning can lead to incorrect answers. 

\begin{figure*}[htbp]
  \centering
  \includegraphics[width=1.0\linewidth]{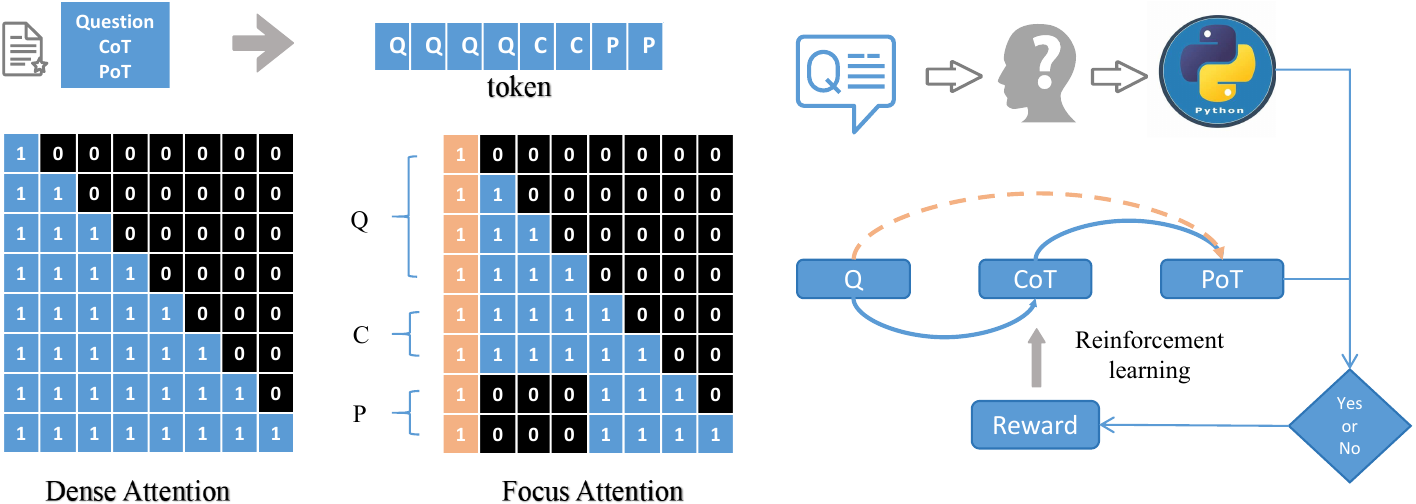}
\caption{Dense Attention refers to traditional Attention, while Focus Attention is our approach. In the orange column on the left, the first four tokens share a consistent mask state of 1. On the right side of the figure, there is a comparison between human and LLMs in solving mathematical problems.}
\label{attention}
\end{figure*}

To more effectively utilize natural language reasoning to enhance PoT, we propose~\textbf{Human-Think Language (HTL)}: A novel information-control-based approach to utilize complete CoT reasoning steps to control PoT generation. HTL is inspired by the way humans write code. Humans consider the entire reasoning process using natural language, and the code can fully rely on natural language reasoning. On the right side of Figure~\ref{attention}, we highlight the parallels between our integrated model and the human approach to solving mathematical problems. Compared to previous works, our framework offers a strong capacity for aligning calculation with reasoning by integrating CoT and PoT. We design Focus Attention mechanism that, during code generation, concentrates solely on information from CoT to promote the chain reasoning better, thereby biasing the answer to be more faithful to CoT. On the other hand, using complete CoT reasoning tends to lead LLMs to use mathematical induction to enumerate reasoning steps verbosely, which results in repetitive generation. We incorporate the error assessment function based on PPO~\cite{ppo}, leveraging reinforcement learning to penalize repetitive generation. We conduct experiments based on CodeLlama-7B and Mistral-7B and achieve outstanding results on eight datasets using only self-distillation data.

In summary, our contributions are as follows:

(1) We are the first to conduct a detailed evaluation of current closed-source models, open-source base models, and specialized models. We highlight the shortcomings of PoT and propose that using full natural language reasoning to enhance PoT performance is essential.

(2) We propose an advanced model named HTL, which utilizes the complete reasoning process of CoT to enhance PoT. HTL incorporates a novel Focus Attention that approximates chain reasoning, complemented by an error assessment function designed to prevent repetitive generation.

(3) We evaluate our work on eight mathematical reasoning datasets, and our experimental results demonstrate that our method achieves outstanding results. HTL shows significant effectiveness in in-domain datasets, out-of-domain datasets, and natural language inference task, demonstrating strong usability and potential.


\section{Method}
The design of HTL is divided into three parts: reasoning format, Focus Attention, and error assessment function based PPO.
\subsection{Reasoning Format}
In Human-Think Language (HTL), we introduce a new reasoning paradigm that uses full CoT reasoning to control the process of PoT, as shown in Figure~\ref{fig:code_with_text}.
The CoT approach, despite sometimes producing erroneous computational outcomes, follows a generally correct skeleton~\cite{wang2024metareasoning} for reasoning.
By adapting code translation after the CoT reasoning path, the PoT can inherit the reasoning skeleton of CoT while circumventing its computational errors. 
This effectively combines the advantages of both approaches.

\begin{figure}[!ht]
\centering
\footnotesize{
\begin{tcolorbox}[
title={\centering \large Human-Think Language}, 
colback=white,
left=0pt,
right=0pt,
bottom=0pt,
top=2pt,
boxrule=0.5pt,
colframe=black!50,
]
\begin{tcolorbox}[
title={\normalsize Question}, 
colback=yellow!10,
left=1pt,right=1pt,top=2pt,bottom=2pt,
]
If $\displaystyle{\frac{2}{5}} = \displaystyle{\frac{A}{60}} = \displaystyle{\frac{60}{B}}$, what is $A + B$?
\end{tcolorbox}

\begin{tcolorbox}[
    title={\normalsize Generated Solution}, 
    colback=blue!10,
    left=1pt,right=1pt,top=1pt,bottom=1pt,
]

We have $\displaystyle{\frac{2}{5}} = \displaystyle{\frac{60}{B}} \Rightarrow B = \displaystyle{\frac{60}{2}} = \textcolor{red!90}{30}$.\\ \\
We also have $\displaystyle{\frac{60}{A}} = \displaystyle{\frac{2}{5}} \Rightarrow A = \textcolor{red!90}{60}$. \\ 
Thus $A+B = \textcolor{red!90}{60+30 = \boxed{90}}$. \\ 
The answer is \textcolor{red!90}{90}. 
\begin{lstlisting}[   
    language=Python, 
    basicstyle=\ttfamily\footnotesize,
    breaklines=true,
    keywordstyle=\color{blue}\bfseries,
    emph={self, print, solve, symbols},
    emphstyle=\bfseries,
    commentstyle=\color{gray}\itshape,
    stringstyle=\color{magenta}
]
# Python program to calculate A + B
from sympy import symbols, solve
# Define the variables
A, B = symbols('A B')
# Equation 1: 2/5 = A/60
eq1 = solve(2/5 - A/60, A)[0]
# Equation 2: 60/B = 2/5
eq2 = solve(60/B - 2/5, B)[0]
# Calculate A + B
result = eq1 + eq2
print(result)
\end{lstlisting}
\end{tcolorbox}

        \end{tcolorbox}
        \caption{Demonstrating a successful example for HTL: Although the CoT's answer may contain many calculation errors (in \textcolor{red!90}{red}), its reasoning skeleton is correct. HTL enables PoT to follow CoT's reasoning steps to arrive at the correct result.}
        \label{fig:code_with_text}
        }
\end{figure}

\subsection{Focus Attention}  

\paragraph{Attention Design}
In our work, we use a local attention~\cite{beltagy2020longformer} mechanism to control the information flow during training (Figure \ref{attention}). We divide the text of a mathematical problem into three parts: $\mathrm{Q}$ (question), $\mathrm{C}$ (CoT), and $\mathrm{P}$ (PoT). The objective in generating the PoT reasoning is for the model to rely solely on information from the CoT reasoning, not on the question. However, a recent study by~\cite{sink} introduced the concept of attention-sink, showing that the initial tokens of a sequence attract a significant portion of attention. Therefore, while the Focus Attention mechanism masks the information from $\mathrm{Q}$ and focuses solely on $\mathrm{C}$ during PoT generation, it preserves the initial tokens in the sequence to prevent the loss of substantial information. Echoing the findings of \cite{sink}, we include the first four tokens in the PoT information generation process. 
This results in the following modified formula for the casual mask matrix:
\begin{equation}
M_{ij} = 
\begin{cases} 
0  & j\leq i \land (j \in \{0,1,2,3\} \lor j \in \mathrm{C}) \\
-\infty & \text{otherwise}
\end{cases}
\end{equation}
Ultimately, the contextualized representation $X^{l}$ of at $l$-th attention layer can be formulated as:
\begin{equation}
\begin{aligned}
A^{l} =\operatorname{Softmax}&\left(\frac{X^{l-1}  W_Q^{l}\left(X^{l-1} W_K^{l}\right)^T}{\sqrt{d / N}}+M\right) \\
X^{l} & =A^{l}\left(X^{l-1} W_V^{l}\right) 
\end{aligned}
\end{equation}
$W_Q$, $W_K$, and $W_V$ are intermediate matrix representations in the attention mechanism. $X$ is the hidden vector representation of the sequence. 
\paragraph{Adaptive Training Strategy}
To align with the dense causal matrix used for both pretraining and inference, which is inconsistent with our Focus Attention, we introduce a novel training approach: during both the initial and final phases of training, we do not explicitly mask any tokens besides the causal mask, thereby ensuring alignment with the pretraining stage and the inference stage. In the middle of the training process, we incorporate a \textbf{mask coverage function}, which is a quadratic function and calculates a proportion of entries to be randomly masked based on the number of training steps, allowing the mask to transition between Dense Attention and Focus Attention:

\begin{equation}
\lambda_{masked}= min(1,-\alpha(\rho_{step}-\frac{1}{2})^2+\beta)
\end{equation}
where $\lambda$ is the percentage of masked entries, and $\rho_{step}$ is the current training step divided by the total steps.
Then, we randomly select the parts to mask in the mask matrix based on the values of the mask coverage function. 
It is noteworthy that during the \textbf{inference phase}, HTL utilizes the traditional causal mask.

\subsection{Error Assessment Function Based PPO} 
In reinforcement-learning stage, We employ PPO with a clipped objective algorithm for training. Following~\cite{ziegler2020finetuning}, the value model $V_\phi$ is constructed by appending a linear value head on top of the last hidden states of the policy model. For the reward model, we replace it with error assessment function. At the terminal state, we use a reward function to score based on the correctness of the generated answer. All other states are assigned a value of 0. We categorize the reasons for their errors and provide more fine-grained feedback scores for the model based on the answers from CoT and PoT. The error assessment function is as follows:
\begin{equation}
\begin{aligned}
f_{r} = \left\{
\begin{array}{ll}
1, & \text{CoT} = \boldsymbol{y}, \text{PoT} = \boldsymbol{y} \\
0.6, & \text{CoT} \neq \boldsymbol{y}, \text{PoT} = \boldsymbol{y} \\
0.3, & \text{CoT} = \boldsymbol{y}, \text{PoT} \neq \boldsymbol{y} \\
0.1, & \text{CoT} = \text{null}\text{ or PoT} = \text{null} \\
0, & \text{CoT} = \text{null}\text{ and PoT} \neq \text{null} \\
\end{array}
\right.
\end{aligned}
\end{equation}

In cases where the model cannot produce an answer, we consider it as model-level error and apply the harshest penalty. If the CoT is correct and PoT is incorrect, we consider it as code translation error. In cases where only CoT is incorrect but PoT is correct, we view it as solely a calculation error and put a slight penalty. Such a partial reward can help reduce the negative effect of learning from sparse rewards. Furthermore, in line with~\cite{DBLP:journals/corr/abs-2307-04964}, our total reward encompasses both the reward function score and the Kullback-Leibler (KL) divergence~\cite{kullback1951information} between the learned RL policy and the initial policy. For the remaining details on PPO, we refer to~\cite{luong2024reft}.

\section{Experiment}

\subsection{Baseline}

Our work is based on the MAmmoTH model~\cite{yue2023MAmmoTH}, which achieves outstanding performance in open-source LLM mathematical reasoning by Hybrid Instruction Tuning from a mixture of CoT and PoT for training\footnote{OpenMathInstruct~\cite{toshniwal2024openmathinstruct} primarily consists of PoT data, which is not suitable for our experimental comparisons.}. MAmmoTH has two bases: CodeLlama-7B~\cite{rozière2023code} and Mistral-7B~\cite{jiang2023mistral}. We compare the following methods:
\paragraph{PoT/PAL~\cite{gao2023pal}} uses the LLM to read natural language problems and generate programs as intermediate reasoning steps, but offloads the solution step to a runtime such as a Python interpreter. PAL is a more refined version of PoT, with each line of code accompanied by a comment.
\paragraph{Hybrid Approach~\cite{yue2023MAmmoTH}} first performs a PoT execution. If the program has any syntax errors, the answer is obtained through CoT.
\paragraph{Rephrase-and-Respond (RAR)~\cite{deng2023rephrase}} enables LLMs to rephrase and expand on the questions posed by humans, followed by the responses within a single prompt. This approach serves as a simple yet effective method for improving performance with a two-stage generation process.
\paragraph{Other Models} such as the strong closed-source model GPT-4 and two 7B open-source models that integrate natural language and code: ToRA~\cite{gou2023tora}\footnote{ToRA perform better on the GSM8K and MATH datasets, but ToRA has not open-sourced its datasets.} and MathCoder~\cite{wang2023mathcoder}.
\subsection{Experimental Setting}
\label{setting}
For reinforcement learning, we set a uniform number of 10 epochs, with the KL coefficient set to 0.01. The learning rates for CodeLlama-Base and Mistral are 1e-5 and 2e-6, respectively. For the SFT stage, the specific parameters are as shown in Table~\ref{set}. For the mask coverage function, we set $\alpha$ to -11.0 and $\beta$ to 1.76. For fair comparison with sota, we follow the standard evaluation protocols\footnote{https://github.com/TIGER-AI-Lab/MAmmoTH}. 
\begin{table}[ht]
\begin{tabular}{ccccc}
\hline
  Model            & Epoch        & Batch Size             & lr               \\ \hline
CodeLlama-Base     & 2           & 64                     & 2e-5           \\
Mistral-Base       & 2          & 64                    & 5e-6               \\ \hline
\end{tabular}
\caption{Details of training hyperparameters for fine-tuning the different base models. Batch size = the batch size per GPU * the number of GPUs * gradient accumulation steps.}
\label{set}
\end{table}

\subsection{Dataset}
\paragraph{Training Dataset} We use hybrid data from the training set of the MAmmoTH model. We first run the fine-tuned MAmmoTH model to generate both CoT and PoT answers for them. We then convert the data into the format of $\mathrm{Q}$, $\mathrm{C}$, and $\mathrm{P}$ and discard any incomplete data. 
In the end, we extract 36,000 examples, with 18,000 coming from GSM8K~\cite{gsm8k}, 3,000 from NumGLUE~\cite{mishra-etal-2022-NumGLUE}, and 15,000 from MATH. Using training data from self-distillation can mitigate the effect of performance differences among models with different bases. 


\paragraph{Test Dataset} Our experiments test on eight datasets: GSM8K, NumGLUE, Math, SimulEq~\cite{koncel-kedziorski-etal-2016-mawps}, DeepMind~\cite{DeepMind}, SVAMP~\cite{SVAMP}, MAWPS~\cite{koncel-kedziorski-etal-2016-mawps} and Asdiv~\cite{miao-etal-2020-diverse}. 
These eight datasets have varying levels of difficulty and length, comprehensively reflecting the model's mathematical computational capabilities. Meanwhile, GSM8K, NumGLUE and MATH are in-domain datasets, whereas others are out-of-domain datasets.

\begin{table*}[ht]
\centering
\resizebox{\linewidth}{!}{
\begin{tabular}{clcccccccc|c}
\hline
Model      &Method             & GSM8K          & NumGLUE &MATH       & SimulEq        & DeepMind                & SVAMP  & MAWPS & ASDiv & Avg       \\ \hline
GPT-4 &PoT &\, 97.0\textsuperscript{*}  &-   &\, 69.7\textsuperscript{*}      & -           & -         &\, 94.8\textsuperscript{*}  &\, 97.7\textsuperscript{*} &\, 92.6\textsuperscript{*}     & -    \\ 
MathCoder &Mix & \, 67.8\textsuperscript{*} & - & \, 30.2\textsuperscript{*} & \, 49.6\textsuperscript{*} & - & \, 70.7\textsuperscript{*} & - & - & - \\ 
ToRA &Tool-integrate & \, 72.6\textsuperscript{*} & 46.2 & \, 44.6\textsuperscript{*} &48.5 & 55.9 & \, 70.4\textsuperscript{*} &\, 91.3\textsuperscript{*} & \, 78.7\textsuperscript{*} & 63.53 \\ \hline
\multicolumn{11}{c}{CodeLlama}   
                   \\ \hline
MAmmoTH &PoT(4-shot)    & 58.6           & 52.5    & 31.7      & 37.4           & 52.0         & 72.1  & 91.7 &68.2     &58.05   \\
MAmmoTH &PAL(4-shot)    & 58.8           & 53.3    & 30.9      & 38.3           & 52.3         & 72.0  & 91.7 &70.3     &58.45   \\
MAmmoTH &PoT    & 58.9           & 56.6    & 32.8      & 44.1           & 53.7         & 70.7  & 91.9 &69.3     &59.75   \\
MAmmoTH &Hybrid & 59.4           & 66.4    & 33.4       & 45.9           & 59.8         & 71.4  & 92.0 &69.3  &62.20\\
MAmmoTH &RAR & 61.2           & 57.3    & 32.7       & 45.3           & 61.2         & 72.1  & 91.6 &72.2  &61.69\\
HTL &Two-stage          & 59.6   & 70.6 & 32.2       & 50.5 & 61.2                 & 70.3    & 92.3 & 70.9 &63.45 \\
HTL &-               & 61.7  & 63.0 & 33.9                 & 48.6           & 61.1          & 71.5   & 92.8 & 71.6  &63.03     \\
&+focus       & 63.9 & 74.1 & 34.1 & \textbf{50.9}           & \textbf{63.1}          & 72.3 &\textbf{95.0} & \textbf{74.0} & 65.96\\
&+RL        & \textbf{68.7} & \textbf{66.9} & \textbf{34.5} & 45.8           &  61.3          & \textbf{76.1} & 92.3 &72.9 &65.81 \\
&+focus+RL        & 65.7 & \textbf{75.1} & \textbf{34.9} & 50.8           &  62.8          & 74.4 & 94.2 &73.1 &66.27 \\ \hline
\multicolumn{11}{c}{Mistral}   
                  \\ \hline
MAmmoTH &PoT    & 74.5          & 73.9     & 37.1     & 48.2           & 55.8        & 80.5  &93.9 &	74.7 &67.33\\
MAmmoTH &Hybrid    & 75.0           & 73.9   & 39.7        & 50.3           & 61.1        & 80.6   &93.9 &74.7  &68.65     \\
MAmmoTH &RAR    & 76.3           & 74.7      & 37.3     & 49.3           & 54.3        & 80.3   &93.7 &74.8   &67.59    \\ 
HTL &-    & 74.7           & 76.3      & 38.5     & 51.6           & 62.9              & 81.2    &93.7 &76.2  &69.38    \\ 
 &+focus    & 77.9           & 77.0   & 39.9       & \textbf{57.8}           & 63.3              & 82.0  &\textbf{94.5} &78.3 &71.34\\ 
 &+focus+RL    & \textbf{78.1}     & \textbf{78.3}  & \textbf{40.6}        & 56.7           & \textbf{64.2}              & \textbf{82.4} &94.2 &\textbf{78.9} &71.67\\ \hline
\end{tabular}
}
\caption{All results are presented as the average of three experimental trials. Results marked as * are copied from other papers. Unless otherwise specified, the default experimental setting is 0-shot. HTL(-) represents that the experiment only used Dense Attention and fine-tuning, while ``focus'' indicates the inclusion of Focus Attention. The ``RL'' is reinforcement learning.}
\label{Table1}
\end{table*}

\subsection{Main Results}
The main results are shown in Table \ref{Table1}. Our method clearly surpasses other existing methods, achieving state-of-the-art (SOTA) across multiple datasets. Most noticeably, our method exhibits a significant improvement on the NumGLUE dataset, because the dataset contains a large amount of natural language inference, which is unsuitable for direct PoT. On average, HTL improved 5\% performance for Llama-Base and 4\% for Mistral-Base. We will discuss some detailed findings below.

For the experiments with PoT (4-shot) and PAL (4-shot), since the current PoT already uses meaningful variable names in the code, adding additional comments by PAL results in a very slight improvement. For the experiment with RAR, while it can reduce some misunderstandings the model has about the problem, it cannot prevent incorrect reasoning. For the hybrid approach that switches to CoT upon errors in code execution, while it has a 9.8\% improvement on NumGLUE over vanilla PoT, HTL achieves an 8.7\% improvement over it by establishing a closer connection between CoT and PoT in a unified one-stage generation process. Compared to proprietary models, GPT-4 still exhibits strong performance, widening the gap with open-source models. ToRA and MathCoder use data generated from GSM8K and MATH datasets. Our performance on these two datasets is not as good as ToRA's, but we have excellent generalizability, showing significant improvements on out-of-domain datasets. Our method of controlling information flow exhibits strong transferability because it directly changes how the model acquires information, making its effectiveness not limited to in-domain datasets.

We also conduct experiments for a two-stage version of HTL and observed no consistent performance gain of vanilla one-stage generation over it. 
This shows that the performance gain mainly comes from Focus Attention and Reinforcement Learning we designed for the one-stage paradigm.   

\begin{table*}[ht]
\centering
\resizebox{\linewidth}{!}{
\begin{tabular}{ccccccccc|c}
\hline
Model        & GSM8K          & NumGLUE   & MATH     & SimulEq        & DeepMind  & SVAMP  & MAWPS & ASDiv  & Avg       \\ \hline
\multicolumn{9}{c}{CoT}   &           \\    \hline
MAmmoTH      & 44.5           & 36.0   & 11.86        &14.7           & 34.2                & 37.0   &75.68 & 60.78 & 39.34     \\
HTL       & 44.1           & 36.2    & 12.1       & 15.03           & 33.8      & 36.9   &75.72 &61.0  &39.37    \\ \hline
\multicolumn{9}{c}{Self Distillation}   &  \\              \hline
MAmmoTH(PoT)  & 58.9           & 56.6    & 32.8      & 44.1           & 53.7         & 70.7  & 91.9 &69.3     &59.75        \\ 
HTL(only-PoT) &  60.6          & 59.6  & 32.7        & 43.7           & 52.7         & 69.7  & 92.0 &71.1  & 60.26\\
HTL         & 61.7  & 63.0 & 33.9                 & 48.6           & 61.1          & 71.5   & 92.8 & 71.6  &63.03  \\ \hline
\end{tabular}
}
\caption{Based on the performance comparison with Llama-Base, to verify the effectiveness of self-distillation in our experiments.}
\label{Table2}
\end{table*}

\subsection{Ablation Study}
We conducted ablation experiments on all datasets to investigate the contribution of each key component or strategy of our proposed method. Ablation experiments include two aspects: method ablation and data ablation. 
\paragraph{Method Ablation} The ablation tests include \textbf{w/o Focus Attention} and \textbf{w/o reinforcement learning}. Focus Attention is a powerful enhancement for HTL, directly improving performance by an average of 2\%. It effectively helps the model focus on useful information. For reinforcement learning (RL), it is noteworthy that both Llama-Base and Mistral-Base models show improvements in math tasks. Math is currently the most challenging mathematical dataset, generating solutions that are often longer than those of other datasets. This frequently causes the model to repeat generation until it exceeds the limit. Reinforcement learning effectively mitigates this issue. But using RL alone significantly improves performance on similar in-domain datasets, but it struggles to transfer to out-of-domain datasets, and its CTE issue remains unresolved.

\paragraph{Data Ablation} We use a self-distillation method to generate data, a technique that has been proven to enhance performance~\cite{zhang2019your}. To demonstrate the effectiveness of our approach, we validate the performance of the HTL model in terms of CoT, and we fine-tune the model using only the PoT subset from the HTL dataset. The results are shown in Table~\ref{Table2}. HTL and MAmmoTH exhibit nearly identical performance in CoT, which is in line with our expectations. Our enhancements predominantly arise from the transition from CoT to PoT, rather than from strengthening the capabilities of CoT. And the data from self-distillation only provide a marginal improvement.

\subsection{Influence of Subsets}

By utilizing training subsets with varying sources and sizes, we can more precisely assess the impact of each data segment on the model's performance. The results show in Table~\ref{subset}. We discover an interesting phenomenon: when we use a specific dataset for downstream training, the model performs well on its corresponding test set but weakens its capabilities on other datasets. 

When we mix multiple datasets for training, the model's improvement in capabilities becomes more comprehensive. The combination of different datasets allows the model to focus more on the characteristics of mathematical problems rather than relying on specific patterns present in only one dataset. At the same time, the addition of GSM8K and NumGLUE has little impact on the MATH dataset; simple mathematical problems are difficult to influence the ability to perform hard reasoning.

\paragraph{Data Volume and Performance Relationship} To explore the appropriate amount of data, we introduced a dataset twice as large for experimentation. The total size of this dataset is 75k, which includes 36,000 entries from GSM8K, 36,000 entries from Math, and 3,000 entries from NumGLUE. As more data is added, the improvement in the model is very slight because we are not injecting more knowledge but rather letting it tend to learn a paradigm. 

\begin{table*}[t]\tiny
\centering
\resizebox{\linewidth}{!}{
\begin{tabular}{cccccc|c}
\hline
              & GSM8K          & NumGLUE    & MATH      & SimulEq             & SVAMP      & Avg   \\ \hline
G             & \textbf{63.3}           & 55.7    & 32.9       & 47.4                    & 71.6   & 54.17 \\
N              & 59.2         & \textbf{64.6}     & 32.4     & 46.6                & 70.7    & 54.7    \\
G/2+N           & 61.2          & 62.3   & 32.8       & 47.1                     & 69.4   & 54.76    \\
G/2+N+M/2       & 62.0         & 63.9    & 33.3      & 48.9                     & 71.9 & 55.5 \\
G+N+M        & 61.7 & 63.0 &\textbf{33.9} & 48.6     & 71.5 & 55.73 \\
2G+N+2M  &63.4 &62.7 &33.7 &\textbf{49.2} &71.8 &56.16\\ \hline
\end{tabular}
}
\caption{G: GSM8K, M: MATH, N: NumGLUE. G/2 indicates that we only utilized half of the generated GSM8K data, with the aim of exploring whether optimal results can be achieved with a lower data volume. 2G indicates twice the data volume of GSM8K.}
\label{subset}
\end{table*}

\subsection{The Effort of Mask Coverage Function}
\begin{table*}[ht]\tiny
\resizebox{\linewidth}{!}{
\begin{tabular}{cccccc}
\hline
    & GSM8K          & NumGLUE   & MATH     & SimulEq             & SVAMP         \\ \hline
Without Inital Tokens     & 13.3           & 7.8     & 1.1      & 3.2                            & 17.8  \\
Mask Coverage=1 & 62.7 & 72.8  &33.6  & 49.6     & 71.9  \\ 
Adaptive Training Strategy    & \textbf{63.9} & \textbf{74.1} &\textbf{34.1} & \textbf{50.9}     & \textbf{72.3} \\ \hline
\end{tabular}
}
\caption{Influence of different Coverage Function.}
\label{relation_of_coverage}
\end{table*}

The phrase ``Without Initial Tokens'' indicates that we block all tokens from $\mathrm{Q}$, not preserving the first four, which significantly decreases model performance, almost rendering it unable to reason correctly. In the second experiment, we set the mask coverage to always be 1, not adapting the model during the initial training phase, nor reverting the attention mechanism to a causal mask during the output phase. In this experiment, we find that its loss convergence rate is significantly slower than the Adaptive Training Strategy. The adaptive training strategy performs better across all datasets, serving as a transitional phase to balance the Focus Attention training mechanism and the inconsistency during inference. We provide the model with a buffer, allowing it to gradually learn local attention, and after training, we restore it to its autoregressive generation mode. The quadratic function accelerates its speed gradually both during ascent and descent and at the peak, the derivative decreases, resulting in a longer duration of focused attention training.

\subsection{Influence of Other Language}
\begin{table}[t]\tiny
\centering
\resizebox{\linewidth}{!}{
\begin{tabular}{clc}
\hline
Model & Method        & Math23k            \\ \hline
\multicolumn{3}{c}{CodeLlama} \\  \hline
MAmmoTH & CoT       & 14.1 \\
MAmmoTH & PoT  &34.7 \\
HTL     &- & 33.6 \\
      &+focus & 30.6 \\
      &+focus+RL & 31.8\\ \hline
\multicolumn{3}{c}{Mistral} \\ \hline
MAmmoTH & CoT       & 28.4 \\
MAmmoTH & PoT  &36.4 \\
HTL     &- & 37.9 \\
      &+focus & 40.2 \\
      &+focus+RL & 40.4\\ \hline

\end{tabular}
}
\caption{HTL in math23k.}
\label{math23k}
\end{table}
We use a Chinese math dataset, math23k~\cite{zhao2020ape210klargescaletemplaterichdataset}, to test the advantages of HTL in other languages. The results shows in Table~\ref{math23k} .We observed that the base model's capability in Chinese CoT constrains the effectiveness of HTL, as HTL relies on effective CoT to enhance PoT. Specifically, when the base model exhibits weak performance in Chinese CoT (e.g., CodeLlama CoT achieves only 14.1\% accuracy), the HTL-family methods perform worse than PoT, potentially because the generated Chinese CoT is ineffective so that it rarely helps but rather undermines PoT. Conversely, when the base model has reasonable Chinese CoT capability (e.g., Mistral CoT achieves 28.4\% accuracy), the HTL-family methods show better performance than PoT.
\section{Analysis}

\subsection{Error Analysis}
\begin{figure}[htbp]
\centering
\includegraphics[width=1.0\linewidth]{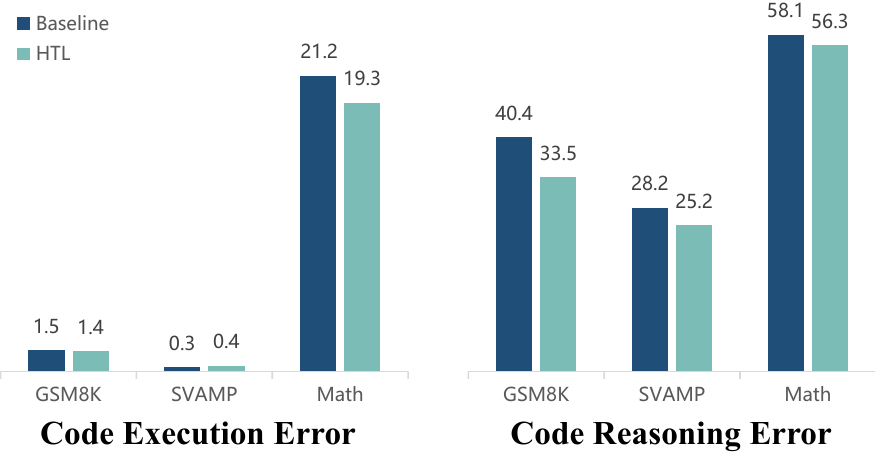}
\caption{Types of errors and their proportions.} \label{error_types}
\end{figure}

To explore how HTL affects model performance and analyze the reasons for errors in various categories, we have divided the errors into two categories: code execution errors and code reasoning errors. Figure~\ref{error_types} shows the proportions of two types of errors across different datasets. For simpler datasets like GSM8K and SVAMP, there are rarely any code execution errors; most are logic reasoning errors, which HTL reduces. For the more challenging dataset like MATH, HTL not only demonstrates stronger logical capabilities but also reduces code execution errors. In HTL, the CTE for CodeLlama-Base and Mistral-Base has been significantly reduced, with CodeLlama-Base decreasing from 8.33\% to 3.96\% and Mistral-Base from 6.97\% to 3.55\%. However, the CTE has not been fully resolved because our data only correlates CoT and PoT based on correctness, not process correspondence. In addition to reducing CTE, part of the performance improvement in HTL comes from correctly solving problems that both CoT and PoT got wrong originally.

\subsection{The Role of Reinforcement Learning}
In experiments, the model enhanced with reinforcement learning shows minimal improvement in average accuracy (only a 0.3\% increase). However, for the MATH dataset, reinforcement learning consistently yields improvements. This improvement stems from reinforcement learning's ability to address the issue of repetitive generation during the CoT in LLMs. When using natural language reasoning, LLMs tend to enumerate answers, leading to repetitive loops until reaching the maximum length limitation. Supervised fine-tuning struggles to suppress this phenomenon, whereas reinforcement learning can effectively penalize it when it occurs.
\section{Discussion}
\paragraph{Do Larger Models Have Issues with PoT?}
~\cite{gao2023pal} achieved good results in testing on LaMDA-137B and PaLM-540B~\cite{anil2023palm} by using text to guide code. ~\cite{wang2023mathcoder} also employed the method of combining natural language with code, which proved effective on a 70 billion parameter open-source model as well. We conduct evaluations on MAmmoTH-Coder-13B and MAmmoTH-Coder-34B, calculating the proportion of \textbf{CTE}. On the five datasets, MAmmoTH-Coder-13B has an average error rate of 8.2\%, while MAmmoTH-Coder-34B has an error rate of 8.7\%. 
CTE does not decrease with the increase in model size. In the future, the amount of training text data will still far exceed that of code data lakes, making it difficult to solve CTE by merely increasing the model size and data volume.

\paragraph{The PoTential of Focused Attention in Other Tasks}
The current autoregressive inference has limitations in that it cannot obtain the solution to a problem before generating the first token~\cite{gloeckle2024better}. CoT can implicitly increase the model's depth, allowing it more extended thinking time to arrive at accurate answers~\cite{NEURIPS2023_dfc310e8}. Extending the model's thinking time to get the right answer before generating the first valid token will be crucial~\cite{goyal2023think}. For reasoning tasks, Focus Attention can gather information and allow large models to concentrate on some intermediate processes (such as setting special tokens) to extend thinking time. On the other hand, Focus Attention can concentrate on the reasoning part of all reasoning tasks while ignoring the question (Q), making the reasoning process more reliable. In several logical/symbolic reasoning tasks, CoT does not significantly outperform directly generating answers~\cite{bao2024llms}. Focus Attention may play a crucial role in these cases.

\section{Related Work}

Current methods primarily rely on the CoT to address mathematical problems, generating reasoning answers in a step-by-step manner~\cite{nye2021work,imani2023mathprompter,miao2023selfcheck,penedo2023refinedweb}. The focus of recent research centers on data engineering and prompt design. In the realm of data engineering, the efforts aim to enhance the quality and increase the volume~\cite{luo2023wizardmath} of CoT data. However, another stream of research identifies several computational challenges associated with exclusively using the CoT approach. In response,~\cite{chen2023program} introduces the PoT, a method that employs Python programs to achieve more accurate results.~\cite{yue2023MAmmoTH} attempts to merge CoT and PoT data in their dataset, resulting in significant improvements. Similarly,~\cite{gao2023pal} seeks to enhance PoT with CoT by weaving CoT snippets into code lines, employing few-shot prompting to guide the model. ToRA~\cite{gou2023tora} uses imitation learning to allow natural language to correct errors in code. MathCoder~\cite{wang2023mathcoder} improves accuracy by closely integrating natural language with code, distilling large amounts of data from GPT-4. OpenmathInstruct~\cite{toshniwal2024openmathinstruct} employs Mistral to explore various problem-solving approaches for each GSM8K and MATH problem, providing a 1.8M open-source dataset to the community.

\section{Conclusion}
In our paper, we identify CTE in mathematical problems and explore how to address the gap between large models' text and code capabilities through text and code interaction. We propose HTL, a method that can more closely integrate CoT and PoT to achieve more accurate reasoning and avoid calculation errors. Our experiment shows that without introducing additional information, our method achieves excellent results merely by controlling the flow of information.

\section*{Limitations}
\paragraph{Lack of Equipment} Due to GPU limitations, our experiments are only conducted on the 7B model, and we did not attempt larger models like the 34B or 70B. Although we provide theoretical feasibility, there is a lack of practical experimental support.

\paragraph{Data Relevance} The CoT and PoT data constructed through automated methods are only associated based on correctness. We still lack human evaluation to determine whether their reasoning processes correspond accurately. This is evident from our experimental results: there is a significant improvement for simpler datasets like GSM8K, as the problem-solving approaches are generally similar. However, for more challenging datasets that may have multiple different solutions, the relevance might be lower.
\paragraph{Exploration of Focus Attention} Regarding Focus Attention, we have not yet determined the specific reason for the need for a gradual increase in coverage to adapt to inference. If we extend this approach to other domains, such as incorporating it into the pre-training stage, it enables the model to better learn step-by-step generation, PoTentially leading to improved results.
\paragraph{Experimental Limitations with Closed-source Models}
We conduct experiments only on open-source large models. Due to the cost of running the models in the API, we do not attempt to explore whether this paradigm can enhance the inference capabilities of models like GPT-4 by constructing similar prompts. However, in our experiments, merely adjusting the prompts does not result in a significant performance improvement. Furthermore, because the training data for GPT-4 is unknown, its results are challenging to interpret.

\bibliography{anthology,main}
\bibliographystyle{acl_natbib}

\newpage
\appendix

\section{Error Case}
\label{sec:appendix}
\begin{figure*}[]
\centering
\includegraphics[width=1.0\textwidth]{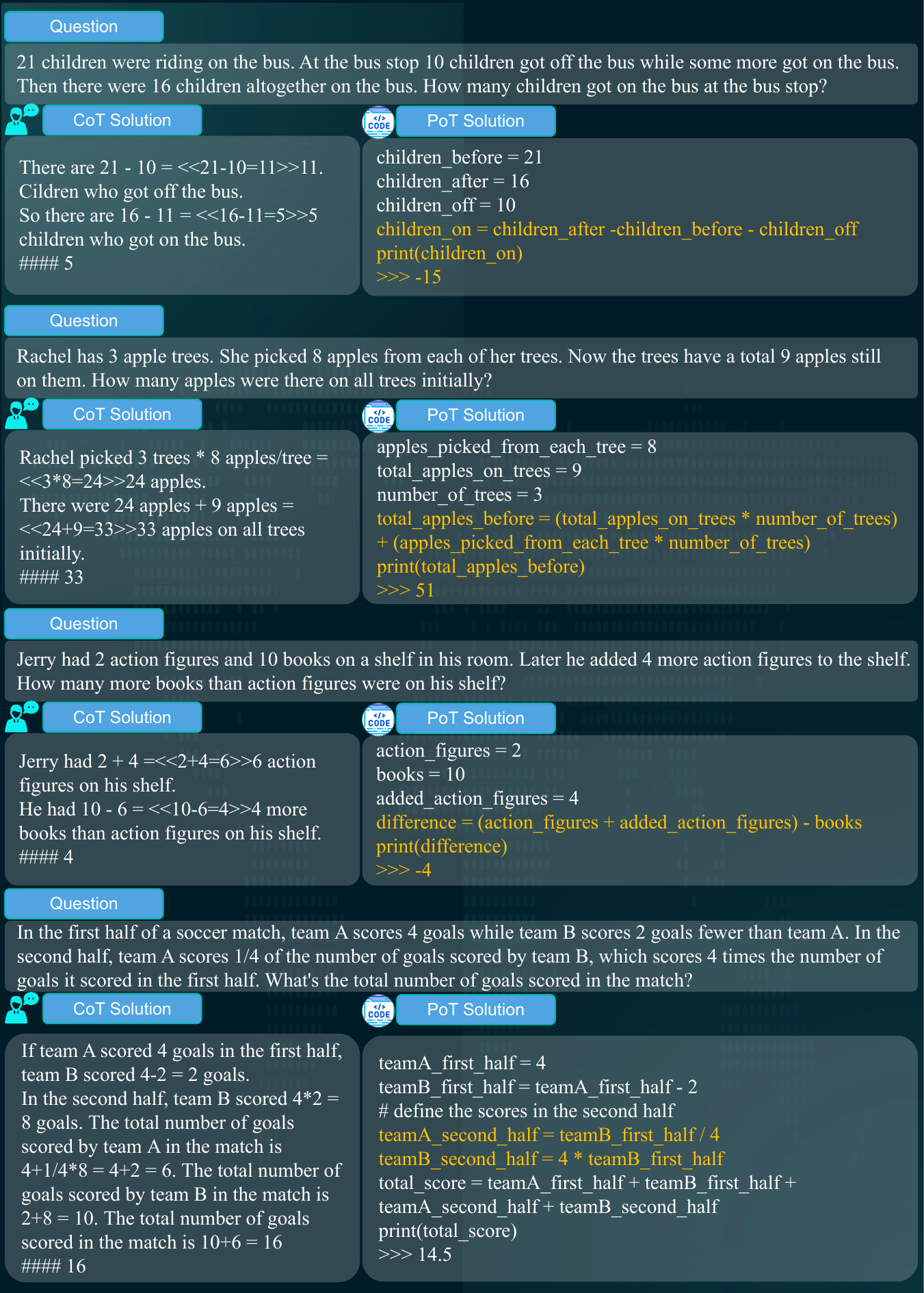}
\caption{The Figure consists of examples where CoT is correct and PoT is incorrect. The first three cases represent logical errors in PoT, which result in incorrect formulas when calculated. The last case represents errors in variable initialization. In PoT answers, such errors typically occur when the initial values need to be computed to obtain the correct result.} \label{error_case}
\end{figure*}

\paragraph{Benchmark Detail}
We conduct measurements from three dimensions: the robust proprietary model GPT-4-turbo, commonly used baseline large models such as CodeLlama and Mistral, and the specialized LLM series MAmmoTH, which are fine-tuned for downstream tasks.
We demonstrate the proportion of instances where models exhibit the phenomenon of being correct with COT but incorrect with PoT in Table~\ref{error_rate}. And we show some examples in Figure~\ref{error_case}.



\begin{table*}[ht]\tiny
\tiny
\centering
\resizebox{\linewidth}{!}{
\begin{tabular}{ccccccc}
\hline
Model       &Base Model             & GSM8K          & NumGLUE        & SimulEq        & DeepMind   & SVAMP  
                  \\ \hline
GPT-3.5-turbo &-    & 11.3          & 18.52          &13.0           & 6.7                & 6.9         \\
GPT-4-turbo &-    & 4.58          & 13.08          &7.55           & 9.64                & 2.1          \\
CodeLlama-7B &Llama-2     & 4.6           & 13.14           &2.3           & 5.5                & 16.8          \\
CodeLlama-7B &Llama-2     & 4.6           & 13.14           &2.3           & 5.5                & 16.8          \\
CodeLlama-13B &Llama-2    & 8.7           & 10.26           &3.6           & 4.8                & 8.5          \\
CodeLlama-34B &Llama-2    & 9.0           & 13.24           &6.6           & 5.5                & 11.5          \\
MAmmoTH-coder-7B  &CodeLlama     & 11.29          & 14.68          & 6.2           & 4.9          & 4.6          \\
MAmmoTH-coder-13B  &CodeLlama         & 9.8          & 16.4           & 6.8 & 3.7                 & 4.0         \\
MAmmoTH-coder-34B  &CodeLlama         & 9.78          & 15.45           & 7.78 & 4.1                 & 6.4        \\
Mistral-7B & Mistral     & 14.1           & 13.5           &10.1           & 9.1                & 9.8      \\ 
MAmmoTH-Mistral-7B & Mistral     & 6.97           & 9.2           &9.1           & 4.3                & 5.3        \\ \hline

\end{tabular}
}
\caption{The detailed CTE values of each model on each dataset. Because the MATH dataset is too difficult for the base model, and MAWPS and ASDiv are too simple for GPT-4, we did not include these three datasets in the figure.}
\label{error_rate}
\end{table*}

\section{Influence of Instruction Length}

\begin{table}[ht]
\begin{tabular}{llllll}
\hline
                     & G              & N              & Si        & d              & Sv         \\ \hline
baseline             & 59.4           & 66.4           & 45.9           & 54.8         & 71.4    \\
tune                 & 60.27          & 73.03          & 48.6           & 61.11        & 71.5         \\
padding              & 59.14          & 63.36          & 44.16           & 52.13           & 69.9         \\ \hline
\end{tabular}
\caption{Padding represents a comparison experiment with the tune experiment, where the input sequence lengths are consistent but without introducing a comparison experiment with CoT answers.}
\label{Table5}
\end{table}

In LLM generation, the length of the input sequence is a crucial factor that provides the model with more time for contemplation. Our experiments utilized a longer prompt template than before. In our test data, the length of most questions is only 20-30 tokens, while the length of our instructions has been extended to 80 tokens. We aim to verify whether the increased input length contributes to performance differences. We replacing all our newly added prompts with special characters such as comma, exclamation mark, and full stop. 
If all prompts and CoT answers replace with a large quantity of the same character, it impact the loss during training, hindering the model from learning information from other tokens. For instance, generating comma for all prompts could result in a satisfactory training loss. The experimental results show in Table \ref{Table5}. 

We find that in mathematical reasoning tasks, simply adding meaningless characters to extend the model's thinking time is ineffective. This highlights the efficacy of the approach proposed by our method.

\section{LLM's Digital Computation Abilities}

The mathematical calculation errors of LLMs originate from two aspects. The first is due to the specificity of how language models understand text—they need to split each sentence into different tokens. For a multi-digit number, it cannot always be treated as a whole. Establishing more advanced token representations is one solution. The other aspect is the internal computational capability for numbers, where multiplication of multi-digit numbers is a significant challenge for LLMs currently.~~\cite{dziri2023faith} shows that GPT-4 achieves only 59\% accuracy on three-digit multiplication problems, while its accuracy for four-digit and five-digit multiplication drops to 4\% and 0\%, respectively.

\section{Why We Use Local Attention}
~\cite{beltagy2020longformer} propose sliding window attention, a mechanism where attention apply in a sliding window manner. It focuses on a fixed-size window around each token, aggregating information within this vicinity for the generation of the current token. In Windows Attention, the source of information generation can be controlled through the attention window. The experiments by~\cite{localattention} and~\cite{pan2023slide} show that local attention mechanisms can focus information more effectively.
\section{Examples}

\label{fig:error_message}
\begin{figure}[!ht]
    \centering

    \footnotesize{

\begin{tcolorbox}[
    title={\centering \large  Max Token Reached}, 
    colback=white,
    left=0pt,
    right=0pt,
    bottom=0pt,
    top=1pt,
    boxrule=0.5pt,
    colframe=black!50,
    ]

\begin{tcolorbox}[
    title={\normalsize Question}, 
    colback=yellow!10,
    left=1pt,right=1pt,top=1pt,bottom=1pt,
    ]
What is the 453rd digit to the right of the decimal point in the decimal expansion of $\frac{6}{13}$.
\end{tcolorbox}

\begin{tcolorbox}[
    title={\normalsize Generated Solution}, 
    colback=red!20,
    left=1pt,right=1pt,top=1pt,bottom=1pt,
    ]
Let $f : \mathbb{R} \to \mathbb{R}$ be a function such that $f(5) = 3$ and
\[f(4xy) = 2y[f(x + y) + f(x - y)]\]for all real numbers $x$ and $y.$  Find $f(2015).$Let's write a program.
<text>Setting $x = 4$ and $y = 1,$ we get
\[f(9) = 2 \cdot 3[f(5) + f(3)] = 18,\]so $f(3) = 9.$
Setting $x = 4$ and $y = 2,$ we get
\[f(12) = 2 \cdot 4[f(7) + f(5)] = 48,\]so $f(7) = 14.$
Setting $x = 4$ and $y = 3,$ we get
\[f(20) = 2 \cdot 4[f(15) + f(11)] = 160,\]so $f(15) = 60.$
Setting $x = 4$ and $y = 5,$ we get
\[f(25) = 2 \cdot 5[f(20) + f(15)] = 500,\]so $f(20) = 25.$
Setting $x = 4$ and $y = 6,$ we get
\[f(35) = 2 \cdot 7[f(30) + f(25)] = 2100,\]so $f(30) = 35.$
Setting $x = 4$ and $y = 8,$ we get
\[f(40) = 2 \cdot 8[f(35) + f(30)] = 4800,\]so $f(35) = 40.$
Setting $x = 4$ and $y = 9,$ we get \\
....
\end{tcolorbox}
\end{tcolorbox}
}
\caption{An example of the repetitive enumeration phenomenon}
\label{error_message}
\end{figure}

\end{document}